\documentclass[conference]{IEEEtran}
\IEEEoverridecommandlockouts
\usepackage{cite}
\usepackage{amsmath,amssymb,amsfonts}
\usepackage{algorithmic}
\usepackage{graphicx}
\usepackage{textcomp}
\usepackage{xcolor}
\def\BibTeX{{\rm B\kern-.05em{\sc i\kern-.025em b}\kern-.08em
    T\kern-.1667em\lower.7ex\hbox{E}\kern-.125emX}}
\begin{document}


\title{Episodic memory in \textsc{ai} agents poses\\risks that should be studied and mitigated}

\author{
\IEEEauthorblockN{Chad DeChant}
\IEEEauthorblockA{Computer Science Department \\
Columbia University\\
New York, NY, 10027, USA\\
chad.dechant@columbia.edu}
}



\maketitle

\begin{abstract}
Most current \textsc{ai} models have little ability to store and later retrieve a record or representation of what they do. In human cognition, episodic memories play an important role in both recall of the past as well as planning for the future. The ability to form and use episodic memories would similarly enable a broad range of improved capabilities in an \textsc{ai} agent that interacts with and takes actions in the world. Researchers have begun directing more attention to developing memory abilities in \textsc{ai} models. It is therefore likely that models with such capability will be become widespread in the near future. This could in some ways contribute to making such \textsc{ai} agents safer by enabling users to better monitor, understand, and control their actions. However, as a new capability with wide applications, we argue that it will also introduce significant new risks that researchers should begin to study and address. We outline these risks and benefits and propose four principles to guide the development of episodic memory capabilities so that these will enhance, rather than undermine, the effort to keep \textsc{ai} safe and trustworthy.

\end{abstract}


\section{Introduction}
\label{introduction}

Among the most significant ways in which current \textsc{ai} models are unlike human cognition is their lack of comparable memory abilities. Very few make any attempt to develop and use an important type of memory on which humans depend, \textit{episodic memory}. Episodic memory is memory of particular past events which one participated in personally and can, in some way, recall \cite{tulving1985elements}. For \textsc{ai} models, this would mean the ability to form and retrieve memories of events --- not merely newly acquired facts --- that happen post-deployment, at runtime.
Most models that incorporate such memory do so in relatively simple ways which are poor approximations of human episodic memory and do not scale to longer, more realistic lengths of time. However, this is beginning to change as more work is done in the area \cite{zhang2024survey, park2023generative}. Making use of episodic memories would enable significant new capabilities and can therefore be expected to be a burgeoning area of research interest in the coming years. 

There is simultaneously a growing interest in developing \textsc{ai} \textit{agents}, models that are trained and equipped to take actions that affect the world \cite{masterman2024landscape, xi2023risepotentiallargelanguage}. Whether these are robots in the real world or virtual agents, they will, by design, be able to impact the world in a more direct way than previous \textsc{ai} models.

\vfill
\noindent\parbox{\textwidth}{\hrulefill\\
\footnotesize Accepted for publication at the 2025 Conference on Secure and Trustworthy Machine Learning (SaTML). The final version will be available on IEEE Xplore.}


While episodic memory plays an important part in many of the cognitive processes which contribute to human intelligence \cite{schacter2016remembering, boyle2022mnemonic}, it has not played a large part in the development of \textsc{ai} agents. This is understandable: until recently, most such agents were limited in the range of actions they could perform and the time horizon over which they could act. They therefore had less need for episodic memories than agents that are the current focus of development.

When \textsc{ai} agents are able to make full use of rich episodic memory abilities, there will be significant implications for their safe deployment. Episodic memories may come to play a role analogous to that which they play in humans, facilitating a wide range of capabilities. These include better planning, problem solving, decision making, and learning \cite{addis2022prospective}. Such capabilities would make any agent equipped with them harder to understand and, in some ways, control.

There is currently an opportunity to prevent episodic memory abilities from making \textsc{ai} agents more dangerous. The implementation and deployment of these abilities are still at an early stage, allowing the research community to study the problem. Possible dangers and benefits of episodic memory can be examined. Most importantly, the results of these studies can be used to guide the implementation of artificial episodic memory to make it safer. There will be a wide variety of ways to implement episodic memory abilities. If the safety implications of these various approaches are understood in advance, research can be directed toward safe techniques and away from more dangerous ones. 

We seek in this paper to draw attention to the risks and benefits of episodic memory in \textsc{ai} agents and motivate a program of research into ways to implement it safely. We begin by summarizing key points of what is known about human episodic memory, including how it differs from other forms of memory more familiar to the artificial intelligence community. In order to show the possible impacts of bringing episodic memory abilities to \textsc{ai} agents, we highlight some of the ways in which humans are thought to use episodic memories, paying particular attention to the many ways episodic memories are used in human cognition beyond simply recalling the past. We also consider whether it is possible for \textsc{ai} to have episodic memories in the way in which humans or, possibly, animals do.



\clearpage

We then outline and explain risks posed by episodic memory abilities in artificial intelligence: deception; retention of knowledge; improved situational awareness; and the unpredictability of memory. We follow this with a discussion of ways in which memories could be used to make \textsc{ai} safer and more trustworthy by facilitating more thorough monitoring, control, and explanations of their actions. We also explain how, counterintuitively, \textsc{ai} agents' own episodic memories may be more easily kept from them than other forms of information, making such memories a crucial element of any effective strategy to control such agents.

In response to the risks we outline, we propose four principles for implementing episodic memory capabilities in a way that promotes, rather than undermines, the safety and reliability of systems with such abilities: 
\begin{enumerate}
    \item Memories should be interpretable by users
    \item Users should be able to add or delete memories
    \item Memories should be in a format which enables users to isolate and detach them from the rest of the system they are part of
    \item Memories should not be editable by \textsc{ai} agents
\end{enumerate}

Finally, we propose some open research areas and questions to encourage further research in this area.

\section{Related work}

\subsection{\textsc{ai} agents}
\textsc{ai} agents have been the subject of study for many decades \cite{wooldridge1994agent}. The most widely used artificial intelligence textbook defines an agent as ``anything that can be viewed as perceiving its environment through sensors and acting upon that environment through actuators'' \cite{russell2016artificial}, where an agent may be a robot operating in the real world or a purely software-based agent operating in a virtual or internet-based environment. \textsc{ai} agents may also be referred to as \textit{autonomous \textsc{ai}}, where autonomous is meant to convey that such systems can ``plan, act in the world, and pursue goals" \cite{bengio2023managing}. 

There are many examples of \textsc{ai} agents in a wide variety of contexts. Robots operating autonomously in the real world are perhaps the prototypical example of such agents as it is easy to see both their independence and tangible effects of their actions. It has been proposed that \textsc{ai} agents operating as biomedical ``\textsc{ai} scientists'' could develop hypotheses, test them in the real world, and have a form of memory to store experimental results \cite{gao2024empowering}.

In recent years there has been an increasing amount of work on \textsc{ai} agents which have a large language model as a component. In these works, \textsc{llm}s are often used to help an agent plan its actions \cite{masterman2024landscape, huang2022language, ahn2022can}, despite evidence that they may not be capable of reliably planning \cite{valmeekam2023planning}. Games have provided environments for the training of agents of various kinds, including  an \textsc{llm}-based agent with a memory capacity for newly acquired skills \cite{wang2023voyager}.

\subsection{Memory in \textsc{ai}}

Techniques patterned after or inspired by episodic memory have been explored in the machine learning literature. This has included work on the efficiency of reinforcement learning through episodic replay \cite{schaul2015prioritized}; planning \cite{eysenbach2019search}; improving world models \cite{codaleveraging}; remembering the values of states or actions \cite{blundell2016model, le2021model}; and more complex memory structures designed to solve tasks which require episodic memories \cite{lampinen2021towards, samsami2024mastering}. Robotics researchers have developed techniques to store and recall information about robots' past actions for use in summarization, question answering, and planning \cite{barmann2022did, dechant2023learning, anwar2024remembr, pointeau2021}. 

Architectures for incorporating different kinds of memory-like functions have led in the past to meaningful improvements in capabilities. These included Long Short Term Memory modules \cite{hochreiter1997long}, Neural Turing Machines \cite{graves2014neural}, Differentiable Neural Computers \cite{graves2016hybrid}, Hopfield networks \cite{hopfield1982neural}, and Modern Hopfield Networks \cite{ramsauer2020hopfield}.

The increasing length of context windows in large language models may raise the question of whether forming representations of episodic memories will ultimately be necessary. Instead, it might be thought that it could eventually be possible to give such a model an agent's entire history as input in a relatively raw format. However, this is unlikely for several reasons. First, it would be very inefficient to reprocess an entire history at every time step an agent acts. Second, the sheer length of time that agents will eventually operate (e.g. decades) would almost certainly be too large for even future long-context models. Third, it possible that such very long context windows will continue to lead to degraded performance, as has been seen in current models \cite{levy2024same}.

The potential for various kinds of memory has received particular attention in the natural language processing community \cite{nematzadeh2020memory}. The need to circumvent a fixed length for input to large language models has inspired many ways of compressing information and storing it for later use by models, including in retrieval augmented generation \cite{wang2024augmenting, zhong2024memorybank, xue2022remembering, wu2020memformer, lewis2020retrieval, guu2020retrieval}.
Moving closer to an agentic framework, recent works have given models greater control over the retrieval and use of information \cite{packer2023memgpt, jiang2023active}.

The existing work which comes closest to our conceptualization of the role of memory introduces a virtual environment of interacting \textsc{llm}-based agents that record and later consult natural language records of their actions, using these to better understand their environment and make coherent and relevant plans \cite{park2023generative}. Partially instantiating our speculation about the utility of past episodes for planning, recent works have developed systems to store and retrieve episodes of action to help guide future decision making in a reinforcement learning context \cite{laskincontext, schmied2024retrievalaugmenteddecisiontransformerexternal}. 

Though a comprehensive overview of different types of memory employed by \textsc{llm}-based \textsc{ai} agents is beyond the scope of this work, a recent paper provides just such a thorough survey \cite{zhang2024survey}.

\subsection{Safe and trustworthy \textsc{ai}}
Concerns about the risks posed by artificial intelligence extend back to its earliest days \cite{10.1093/oso/9780198250791.003.0018}. Research on various kinds of harms that artificial intelligence might cause is now the subject of several research communities with a variety of interests \cite{russell2015research, birhane2022values, barocas2023fairness, weidinger2021ethical}. We focus here on work that most closely concerns risks which could be amplified or reduced by episodic memory.

Given the potential of agentic \textsc{ai} to take actions in and affect the world, special attention to the possible harms of \textsc{ai} agents is warranted. There is, therefore, a growing body of research on the possible dangers of such systems and techniques for ensuring they operate safely and understandably \cite{anjomshoae2019explainable}. Attention has been given to developing ways to make robots' actions explainable \cite{sakai2022explainable}, for example in the case of autonomous vehicles \cite{chen2023driving}. \textsc{llm}-based agents, including those operating on the internet \cite{nakano2021webgpt}, pose several ethical and safety challenges which are actively being studied. Agents' use of tools is the focus of a framework that pairs emulation and evaluation of \textsc{llm}-based agents \cite{ruan2023identifying}. A recent paper surveying the risks potentially posed by \textsc{llm}s includes a section on the particular risks of \textsc{llm}-based agents \cite{anwar2024foundational}.

Researchers have given particular consideration to the risks particular to agentic \textsc{ai} that can act autonomously in pursuit of goals its users do not intend, either because of a malicious actor or by accident \cite{bengio2023managing}.  It has been proposed that due to reward misspecification during training, an \textsc{llm}-based agent trained using deep learning and reinforcement learning from human feedback might develop undesirable internal goals, particularly if such agents can determine if they are currently being trained, evaluated, or deployed and operate over long time horizons\cite{ngo2022alignment}. \textsc{ai} agents have been found by many researchers to engage in various kinds of deception \cite{park2024ai}.

Episodic memories could be used to invade a user's privacy. Ways in which machine learning can violate users' expectations of privacy have been extensively studied \cite{liu2021machine}, along with approaches to mitigate these invasions of privacy \cite{abadi2016deep}. Of particular relevance is work which examines the potential for mass surveillance which modern machine learning techniques afford and which widespread agents with episodic memories might contribute to \cite{kalluri2023surveillance}. Dialog agents might have a range of negative effects on their users, including invasions of privacy \cite{ruane2019conversational}.

\section{Episodic memory}
In this section we present a brief overview of episodic memory in humans. Our goal is to provide a short introduction to the topic and to highlight ways in which implementing episodic memory could be useful in creating more capable artificial intelligence. We pay particular attention to the links between episodic memory and other abilities, at least in humans, as these are potentially surprising and are especially significant for considering the effects of memory abilities on \textsc{ai} agents.

\subsection{Taxonomy of memory types}

\textit{Episodic} memory in humans is memory for events in which someone personally participated. The psychologist Endel Tulving is recognized as being the first to propose a distinction between episodic memory and \textit{semantic} memory, which is memory of facts about events and the world \cite{tulving1972episodic}. For example, someone remembering a trip to Paris that they took a few years earlier would be using their episodic memory, while someone remembering that Paris is the capital of France would be using semantic memory. Although semantic memory is typically about impersonal information such as geographical knowledge, it might also be about factual information about oneself which does not call to mind a particular associated episode. For example, remembering which city one was born in would be considered an example of personal semantic memory. 

Both episodic and semantic memory are referred to as types of declarative memory 
\cite{squire1982neuropsychology}. Their contents can (to some extent) typically be described using language. A third kind of memory, \textit{procedural} memory, is sometimes included in taxonomies of memory types and is considered nondeclarative. Procedural memory is memory of how to do something, such as riding a bicycle or other skill or ability which had to be learned \cite{squire2011cognitive}.

Much --- indeed, perhaps most --- machine learning research involves either what we have just described as semantic or procedural memory, though these are not typically described as forms of memory in a machine learning context. Large language models are valued in large part for their semantic memory of facts about the world. This kind of memory has, for example, recently been investigated in the many papers asking what \textsc{llm}s ``know” \cite{burns2022discovering, roberts2020much}. 
Procedural memory for learned skills and abilities is the objective of much machine learning work such as that on learning navigation, game playing, automobile driving, robotic manipulation, etc. 



\subsection{The stages of episodic memory}
While there are many theories and debates about the way in which episodic memories are formed and maintained in the brain \cite{amer2023neural}, in broad outline the process consists of the following stages:
\begin{enumerate}
    \item \textit{Encoding:} Raw sensory and other (e.g. emotional) information about an episode needs to be compressed and structured into a suitable representation. Forming representations of episodic memories critically depends on the hippocampus.and nearby structures in the medial temporal lobe. Damage to the hippocampus is known to impair the ability to store new episodic but not semantic memories \cite{vargha1997differential}.\\
    Episodic memories do not depend only on the hippocampus, however. According to the hippocampal indexing theory \cite{teyler1986hippocampal}, an encoding of an episode in the hippocampus serves as a kind of index that points to and binds together representations in the neocortex that form the basis of the episode, such as multimodal sensory representations along with associated emotional and conceptual information \cite{moscovitch2016episodic}.
    \item \textit{Storage:} After a memory is encoded, it undergoes a period of so-called \textit{consolidation} or transformation into a long term form \cite{dudai2012restless}. This is commonly thought to involve moving the representation from the hippocampus to the neocortex, either partly \cite{nadel2000multiple} or entirely \cite{squire1982neuropsychology}.\\
    The memory must then \textit{persist} in its stored form \cite{persistence}. There is evidence that a memory is in some way destabilized when it is retrieved or activated, leading it to undergo \textit{reconsolidation} \cite{nader2000fear}.

    \item \textit{Retrieval:} Retrieval is the recovery of a previously encoded episodic memory \cite{gardiner2007}. A \textit{cue}, which could be an activation of part of the pattern that was stored in memory, triggers the associated hippocampal index representation. This index is then thought to activate the associated neurons in the neocortex which made up the original encoding of the episode \cite{teyler2007hippocampal}.
\end{enumerate}

From this necessarily quite abbreviated account, we may draw a few conclusions relevant to the consideration of episodic memory in artificial intelligence. First, episodic memories depend on many brain regions for their creation and persistence. Episodic memories are not located in just one area that is easy to study and manipulate. Second, humans' episodic memories are not simply left untouched after their creation; they are neither static nor unchanging. Third, there would be many points of possible intervention in an artificial process that mimics the above stages. Such forms of intervention could be used to enact the recommendations for safe episodic memory proposed later in this paper.

\subsection{The uses of episodic memory}
Episodic memories are thought to be involved in a variety of important cognitive processes beyond simply recalling past events. Evidence for this comes from two primary sources. First, brain imaging studies have shown that similar brain regions are recruited during recall of episodic memories and other tasks \cite{szpunar2007neural}. Second, people with impairments in episodic memory abilities are found to also have deficiencies when performing other tasks \cite{klein2002memory}. These two kinds of evidence, along with theoretical accounts that seek to explain the observed relationships, suggest that episodic memories --- or at least the ability to form episodic memories and its associated cognitive architecture --- are used when performing many other important cognitive functions.

Episodic memories are, as memories, naturally of events in the past. They are used, however, to influence the future. Indeed, some have argued on evolutionary grounds that memory should be considered to be primarily concerned with the future, helping us act in whatever new circumstances the future presents us \cite{klein2013temporal}.

\textbf{Planning} Especially relevant to the concerns of this paper is the way memories are used when planning future actions: according to some theories, memories serve as ``building blocks", allowing elements of particular episodes to be reused and reassembled in different ways in order to respond to novel situations \cite{schacter2007remembering}. Some psychologists have gone so far as to suggest that ``episodic reconstruction is just an adaptive feature of the future planning system" \cite{suddendorf2005making}.

\textbf{Imagination and prediction} 
Accumulating evidence shows that episodic memory --- and the brain systems that support it --- is involved in predicting and imagining the future. Patients with damage to hippocampal and non-hippocampal regions involved in memory have unusually poor performance when asked to predict or imagine future scenarios. Such patients imagine impoverished scenarios lacking in detail and coherence \cite{buckner2010role, hassabis2007patients}.

\textbf{Problem solving} As long ago as the nineteenth century it was observed that patients with amnesia lacked the ability to engage in flexible thinking, with one doctor observing of his amnesiac patients that the ``circle of ideas in which the patient's intelligence moves becomes very restricted" \cite{korsakoff1889etude, buckner2010role}. An association has been found between having deficits in episodic memory functioning and being unable to generate relevant details in an open-ended problem solving task \cite{sheldon2011episodic}.

\textbf{Decision making} 
One proposed psychological model demonstrates how episodic memories can help in learning a new task by allowing successful episodes to be recalled and emulated \cite{lengyel2007hippocampal}. This approach was extended to show how similar memories could be sampled in order to estimate the value of possible actions \cite{gershman2017reinforcement}. 

\textbf{Learning from episodic memories}
Episodic memory has been described as ``epistemically generative" in the sense that it enables learning from past experiences \cite{boyle2019learning}. Past events may, for example, be recalled and reinterpreted in light of newly acquired information, allowing one to learn from remembered aspects of the past events which had previously been misunderstood.


\subsection{Can an \textsc{ai} agent have episodic memories?} The wide variety of ways episodic memories are used by humans suggests that the incorporation of true episodic memory abilities into \textsc{ai} agents would greatly expand their range of capabilities. But some may question whether it is even possible for \textsc{ai} agents to have actual episodic memories. Indeed, Endel Tulving himself described the phenomenon of recalling episodic memories in terms which cast doubt on the very idea. He wrote that remembering a past episode is a kind of ``mental time travel,'' a ``conscious awareness of what had happened in the past'' which has an ``experiential flavor" \cite{tulving2002episodic}. 

Some psychologists and philosophers have concluded that episodic memory is therefore a uniquely human phenomenon, lacking even in non-human animals --- much less \textsc{ai} agents. According to this view, animals are ``stuck in time,'' without either episodic memory or ``the ability to anticipate long-range future events'' which we have seen is associated with episodic memory \cite{roberts2002animals}. 

Others, however, have taken a more expansive view of what constitutes episodic memory as well as who has it. Rather than focusing on phenomenological aspects of memory, it is possible instead to consider episodic memory as simply combining memory for \textit{what} happened, \textit{when} it happened, and \textit{where} it happened \cite{griffiths1999episodic}. While we do not know what non-human animals experience, we can study their behavior. Some studies have indeed found evidence for episodic memory in animals. For example, an experiment demonstrated that birds (scrub jays) reliably behaved as if they remembered what kind of food they had hidden, where they had hidden it, and when they had hidden it \cite{clayton1999scrub}. Another study on scrub jays demonstrated that they also appear to plan for the future when hiding food \cite{raby2007planning}. Others argue that episodic memory has a long evolutionary history that predates humans and point to brain areas in other mammals as well as birds which they claim correspond to areas in the human brain responsible for episodic memory \cite{allen2013evolution}.

If non-human animals do have episodic memory, that would reinforce the concerns which motivate this paper. If it turns out that episodic memories are necessary to achieve a level of intelligence on par with such animals, it would be much more likely that artificial intelligence researchers will conclude that similar memory abilities are needed in \textsc{ai} systems, including agents. Of course, we have seen that researchers are not waiting for this debate to be resolved to move ahead with attempts to engineer episodic memory into \textsc{ai} agents.

Despite the concerns mentioned at the beginning of this section, we suggest that it does make sense to think about \textsc{ai} agents' having episodic memory abilities, at least at the level of animals. It is likely that that researchers will develop architectures that enable combining representations of what, when, and where an agent performed certain actions or witnessed certain events, thus endowing \textsc{ai} agents with at least the functional equivalent of episodic memory.

\section{Risks of Episodic Memory}

Equipping an \textsc{ai} system with episodic memory will allow such a system to operate in new and different ways, some of which present novel risks. These risks include:


\subsection{Deception} Episodic memories can be used to enable an agent to engage in sophisticated forms of deception. It is of course true that one does not need to have episodic memory in order to attempt to deceive others. For example, simply having a policy of always denying that an undesirable action occurred or is planned for the future is a simple form of deception which requires no access to relevant memories or plans (e.g. ``I did not do that”, ``I will not do that”). 

However, more complex forms of deception would be difficult or impossible to carry out without some kind of episodic memory. If an agent is to execute a multi-stage plan over an extended period of time, the agent will have to keep track of both what it has done as well as what it is has already reported to others about its actions in order to maintain an effective deception. 

There is already some evidence to support this concern about deception. For example, one experiment \cite{scheurer2024large} provided an \textsc{llm} (\textsc{gpt-4}) with a simple text scratchpad to record its ``chain of thought" \cite{wei2022chain} reasoning, which functions as a crude form of memory. When ``pressured" to perform an illegal act in a simulation, an \textsc{llm} \textit{with} a scratchpad was found to engage in ``strategic deception" \cite{park2024ai} approximately three times as often as the same \textsc{llm} \textit{without} a scratchpad.

\subsection{Unwanted retention of knowledge} An \textsc{ai} agent equipped with episodic memory might remember things its user would prefer that it not remember. It could then share that knowledge with people or organizations its user does not want to share it with, possibly constituting significant risks to the user's privacy or personal safety. Invasions of privacy are likely to occur in several domains:
\begin{enumerate}
    \item\textit{Interpersonal:} One person could use an \textsc{ai} agent to spy on another. For example, someone could use a household robot's episodic memories to covertly monitor other members of the household.
    \item\textit{Commercial:} Corporations could use \textsc{ai} agents, particularly those they sell or rent, to gather commercially valuable information about their users.
    \item\textit{Governmental:} Governments, especially but not exclusively authoritarian ones, might demand or secretly access the memories of \textsc{ai} agents looking for evidence of forbidden political organizing or the expression of unwanted views.
\end{enumerate}

\subsection{The unpredictability of memory}
We summarized research into humans' use of episodic memories in thinking about the future in Section III(C). If, as we discussed there, episodic memories are used to form the building blocks of future action plans, \textsc{ai} agents might engage in complicated and unpredictable behaviors as a result. This unpredictability derives from two sources: the unknown sources of memories and the difficulty of foreseeing how memories might be utilized.

\textbf{Unpredictable sources of memories}
An \textsc{ai} agent with the ability to form episodic memories will in the course of its operation store many memories that record the various actions it takes and events it participates in. Because these events will themselves be influenced by the actions of humans and, perhaps, other \textsc{ai} agents, what constitutes the stock of memories an agent will come to have must be unknowable before the agent is deployed, acts and, in so doing, creates memories. It will also be constantly changing. Without a significant effort to put limitations on the characteristics of new memories which may be allowed to affect an agent's actions, their influence would be unpredictable.

\textbf{Unpredictable uses of memories}
As we reviewed earlier, humans make extensive use of their episodic memories to understand and act in new situations. If \textsc{ai} agents come to have this ability, the ways that they use that memory will be similarly hard to predict.

Users may be surprised by how such agents use their memories. They may, for example, remember the location of objects that they then use when the user would prefer that they not use them. An agent may participate in a complex action episode while not fully understanding what is happening in the episode; if it later tried to use that episode as an example to draw on when planning a new sequence of actions, its faulty understanding may lead to undesirable and unexpected results. A household robot may, for example, observe one instance of its owner going over to the next door neighbor's apartment to borrow some sugar and then try to do the same when it is asked to bake cookies, not understanding that the asking and receiving of permission from the neighbor is a prerequisite of entering their apartment.

Our hypothesis that episodic memories could be recalled and used in ways which lead to unpredictable and potentially undesirable agent behavior has a strong parallel in existing work on the effects of examples given to large language models. A significant part of the success of \textsc{llms} is their ability to learn new tasks by being given even a few examples as context in their prompts \cite{brown2020language}. Several research groups have demonstrated that such few-shot in context learning presents opportunities to undermine or defeat elements of the models' training which are meant to keep them aligned to particular values such as being harmless \cite{rao2024tricking, wei2023jailbreak}. Recent work has also shown that the use of a large number (e.g. several hundred) of examples has even greater potential to override \textsc{llm}s' safeguards \cite{anil2024manyshot}. We suggest that a set of recalled episodic memories, assembled on the fly as needed and functioning in a way analagous to in context examples, could be a source of similar so-called ``jail-breaking.'' Such a collection of episodes with the ability to negatively influence \textsc{llm} outputs might be assembled accidentally or through some intentional effort on the part of a bad actor. It has been argued that their vulnerability to dangerous in context examples may be an inherent limitation of transformer-based \textsc{llms} \cite{milliere2023alignment}; if so, such a vulnerability could be equally difficult to overcome in similar systems augmented with episodic memory abilities.

\subsection{Improved situational awareness} 

Many contemporary large language models have a great deal of knowledge of the world in general \cite{chang2024largelanguagemodelsacquire}; they can somewhat reliably (hallucinations \cite{ji2023survey} notwithstanding) answer factual questions about information that was in their training data. However, they have little understanding of their own particular circumstances outside of whatever prompt they may be given. Researchers have employed the term ``situational awareness'' to refer to a model's ability to connect the general information it has about the world with the details of its own particular circumstances \cite{cotra2021without}. Some have employed the term in a more restricted sense, to refer specifically to a model's knowledge of whether at any given time it is deployed or undergoing testing (e.g. to test its truthfulness or harmlessness) \cite{shevlane2023model}. Models that have better situational awareness in the general sense might be more broadly capable and better able to take actions which affect the world while those with knowledge of their own training status might evade safety tests by responding in approved ways during training and testing but going on to act in undesirable ways during deployment.

Several experiments using purpose-built datasets have been conducted to test large language models' levels of situational awareness \cite{laine2023towards}. Some of these experiments have shown that current models have only weak situational awareness \cite{perez-etal-2023-discovering} while others suggest the level might be higher \cite{ngo2022alignment}, although it is unclear how much of what appears to be awareness of its own situation might be derived from either its prompt or a combination of its prompt and  information about \textsc{llms} in general.

We propose that a model without episodic memory can have only a very limited form of situational awareness. With no understanding of what actions it has taken in the distant or recent past, what environments it has seen or tasks it has completed, an agent could not be said to have much awareness of its situation. Endowing it with episodic memories would allow it to develop a better, more complete picture of the world and its role in it, allowing for more effective planning and action taking to influence the environment and to achieve objectives. It could use its episodic memories to build up an understanding of those with whom it interacts, the kinds of tasks it performs, and the contexts in which it performs them. It would also develop a knowledge of its own capabilities and limitations that can in some cases only come from observing and later recalling one's own actions, successes, and failures.

Ideally, this would simply lead to an agent better able to perform the tasks its users assign it. But without some check, this improved awareness could represent an enhanced danger in a misaligned agent or one under the direction of a bad actor. For example, episodic memories could allow an agent to learn regularities in the timing or content of safety audits which might be performed either before or during deployment, and thus to evade them. 

\section{Safety benefits of episodic memory}

In contrast to the concerns elaborated upon above, episodic memory could also be used to make \textsc{ai} safer in multiple ways:

\subsection{Monitoring} We cannot ensure that \textsc{ai} operates safely unless we know what it is doing. As \textsc{ai} agents become more capable, they will increasingly operate outside of direct human supervision. Robots may undertake long and complicated tasks that take them far away from their operators; non-embodied \textsc{ai} agents may direct and supervise the operation of complicated systems such as power grids or engage in virtual consultations with humans over medical or legal matters. In these cases and many others it will be impractical or impossible for any human to watch everything that such an \textsc{ai} agent does. It will instead be necessary to rely on \textsc{ai} agents to remember, recall, and share information about their actions.

Several methods were recently proposed to achieve ``visibility into \textsc{ai} agents," one of which was \textit{activity logging} \cite{chan2024visibility}. Episodic memories could be used one way to achieve such logging, as well as to address other calls for research into ``scalable oversight" \cite{amodei2016concrete} and ``monitoring" \cite{hendrycks2021unsolved}. Artificial episodic memory representations could, though, be structured to be more useful and accessible than simple logging.

\subsection{Control}
Maintaining and sharing episodic memories with an appropriate authority could be used as a means to ensure an \textsc{ai} agent is operating as intended and is therefore under control. It could, for example, help prevent the misuse of dual-use technology. Dual-use technology, which can be used for civilian or military purposes, is a particularly significant problem for artificial intelligence, given the general purpose nature of much current machine learning research. It has been plausibly claimed that most \textsc{ai} is dual-use \cite{de2020ethics}.

Having access to an \textsc{ai} system's memories would be one way for a corporation or government to ensure it was not being used in violation of an understanding that it only be used for peaceful purposes, perhaps as part of an export control regime \cite{brockmann2022applying}. Given the high risks associated with the weaponization of \textsc{ai} \cite{simmons2024ai}, techniques and frameworks could be developed to use episodic memories of potentially dual-use \textsc{ai} agents to maintain control over them.

If systems are developed that explicitly make use of episodic memories as building blocks for planning actions, new avenues for control would be opened up. As we will discuss further in Section VI(B), an agent's collection of memories could be curated in order to shape its future actions.

\subsection{Explainability} An accurate history of \emph{what} an agent did is a prerequisite for trying to explain \emph{why} it acted as it did. Thorough memories should include both information about an agent’s perceptions of the environment as well as some record of how its internal states, such as goal representations, interacted with those perceptions to lead to specific actions.

\subsection{Uniquely controllable type of information}
Several aspects of \textsc{llm}-based agents make it difficult to control what information, or even skills, they may have after deployment. First, although there is a great deal of research effort going into deleting information from their weights after they are trained \cite{mitchell2022fast, meng2022locating, meng2023memit}, it is not yet clear how to do this reliably. Information that was thought to be deleted may, in some circumstances, be recoverable \cite{patilcan, łucki2024adversarialperspectivemachineunlearning}. 

Second, and more significantly, given access to the internet, \textsc{llm}-based \textsc{ai} agents could find anything available there, potentially giving them access to information that was deliberately excluded from their training data. This could include examples of skills or behaviors which the agent was not trained on but which it could learn through one- or few-shot incorporation into its context window. \textsc{ai} agents are able to both search the internet \cite{webgpt2022, webcpm2023} and learn new skills in this manner today \cite{brown2020language, li-etal-2024-language}. 

Any publicly available information about the world in general and about skills an agent might acquire will therefore be difficult to keep from an \textsc{ai} agent. By contrast, information about an agent itself and its own unique history will not be widely available. If episodic memories about an agent's past actions are stored, controlled, and managed in the ways suggested in this paper, information about an agent and its own past would be the easiest to selectively keep from it. 

\section{Principles for enabling safe\\and trustworthy episodic memory}

We suggest the following four principles to guide research and implementation of episodic memory abilities in \textsc{ai}:

\subsection{Interpretability of memories} Memories should be accurately interpretable by humans, either directly or indirectly. \emph{Directly} interpretable memories would be in a readily understandable form such as video, images, or natural language. It might in some limited cases be possible to equip an \textsc{ai} agent with useful memory which consists entirely of records in such formats by, for example, recording raw video before it is processed through a vision system. 

It is likely, however, that memory records entirely in such raw formats (especially video) would be impractical; they might be excessively large and difficult to search, access, and make use of. In practice memories are likely to be compressed into smaller representations which would then need to be \emph{indirectly} interpretable. Memories might be indirectly but still reliably interpretable if the memories could yield accurate information which is complete and relevant to a user’s specific interests in monitoring them. A memory might be summarized in natural language \cite{dechant2023learning}, giving the most important events which took place in a given episode; systems could be trained to produce safety-specific summaries, reporting only actions which could be dangerous or otherwise raise concerns about an agent's reliability.

Memories should also be usable for question answering. If a user wants to know something specific about an episode, perhaps in response to a summary, memories should be queryable in natural language. Such queries should not be limited to one episode at a time; memories ought to be able to be compared to other memories, allowing such questions as, “what was different this time?” Memories should also be easily searchable using natural language, allowing users to ask if a particular agent has ever done something, or when something was done, for example. Finally, if the method of compressed representation of episodes allows it, memories might to some extent be visualizable. 

In addition to the above methods for users to be able to interpret memories, techniques from the growing field of research on the interpretability of machine learning methods \cite{ casper2022sok, nanda2023progress} can be applied to the memory representations and help to guide the development of such representations to be intelligible and controllable.

\subsection{Addition or deletion of memories} Users should have complete control over the memories retained by an \textsc{ai} agent. Most importantly, a user should be able to delete memories of particular episodes. A user might not want an agent to remember something for a variety of reasons, from safety-related concerns to more mundane issues, including concerns about privacy or maintenance of trade or government secrets. Conversely, it might be useful for users to add memories of episodes which a particular agent did not itself experience to its store of memories.

The addition or deletion of memories might be particularly important if, as discussed above, \textsc{ai} agents will be able to use and recompose memories to construct new plans for future action. Such episodes might be positive examples of action sequences which a user wishes an agent to repeat or draw upon to incorporate in future plans. Alternatively, it may be useful to give agents memory-like records of episodes which represent undesirable actions; such episodes could function as a kind of warning to allow agents to recognize if they are beginning to carry out actions which are similar to those in an episode added to the agent in order to serve as a negative example. In other cases it might be better for agents not to remember things which their users do not want them to be able to repeat or call upon when planning. 

If agents make use of their memories when planning actions, the addition or deletion of memories could help produce either standardized or specialized agents. In some circumstances it might be best for all agents to have the same stock of memories which might influence their actions, helping to ensure that their behavior is predictable and regular. In others cases, there may be a need for particular agents to maintain their own memories which are never shared, in order to prevent the spread of potentially dangerous information.

\subsection{Detachable and isolatable memory format}
\label{detachable_and_isolatable_format}

Memories must therefore be in a format which allows for their addition or complete deletion by users. This will impose some design constraints on how episodic memories are instantiated in an \textsc{ai} agent because they will have to be in a format which can be cleanly separated from the rest of the system’s architecture. As we saw earlier, the mechanics of human memory are much messier: although some areas (notably, the hippocampus) are more centrally involved in human memory formation and retrieval than others, complete episodic memories are thought to be composed of elements distributed in many areas of the brain \cite{teyler2007hippocampal}. According to some theories of memory, regions with a relative specialization in particular modalities (e.g. vision) are also responsible for storing their respective modality-specific components of a particular memory \cite{amer2023neural}.

Memories which are tightly integrated with and spread throughout many areas would be difficult to delete or add to, so it is likely that memory will have to be designed very differently in \textsc{ai} systems than it is in humans if it is to be implemented in accordance with these safety-oriented principles. This might mean that some of the ways in which humans are able to use memories effectively would not be directly translatable to artificial intelligence, thereby limiting such artificial capabilities relative to those in humans. However, alternative implementations of episodic memory which conform to the above principles may be invented which would allow for memory capabilities which are both safe and effective.

The potential difficulty of removing memories that are tightly integrated into an \textsc{llm} is demonstrated by the body of work done on ``machine unlearning'' \cite{cao2015towards, bourtoule2021machine, mitchell2022fast, meng2022locating, meng2023memit, li2024wmdp}. Techniques to remove information from \textsc{llm}s have been found to be not yet consistently reliable \cite{patilcan, łucki2024adversarialperspectivemachineunlearning}. It is important to note that these techniques are currently not directly applicable to the problem of removing episodic memory because they focus on deleting information found in \textsc{llm}'s weights as a result of its training. By contrast, episodic memories would only be accumulated and stored after training is complete. If future attempts to retain episodic memories involve storing them in a distributed fashion throughout an \textsc{llm}, these techniques might be useful, though only if they develop to be more reliable. 

The most straightforward method for storing and retrieving representations of episodic memories for \textsc{llm}-based agents would likely be similar to retrieval augmented generation (\textsc{rag}) systems. Indeed, a recent \textsc{llm}-based system with an implementation of episodic memory-like abilities for episode sub-trajectories employs \textsc{rag} \cite{schmied2024retrievalaugmenteddecisiontransformerexternal}.

\subsection{Memories not editable by \textsc{ai} agents} In contrast to — and in some tension with — the principle that memories should be able to be easily added or deleted by users is the countervailing principle that memories should not be editable by \textsc{ai} agents themselves. Although memories will have to `edited' when they are created, they should afterwards be left intact and unaltered by the agent. This is necessary in order to ensure that memories remain accurate and uncorrupted. An \textsc{ai} agent should not be able to add, delete, or change its memories. Otherwise, a memory-facilitated form of reward hacking \cite{amodei2016concrete} might occur: if a reinforcement learning-based agent’s reward were tied to a measure of performance which it reports using its memory, it might find that it can achieve a higher reward by altering its memory of its actions rather than by changing what actions it takes.


\section{Safe episodic memory: research questions}

Because episodic memory in \textsc{ai} agents is still at an early stage of development, there is a great deal of work to do to better understand how it will affect agents' abilities in practice and what its effects will be from a trust and safety perspective. Building off of the preceding discussion of risks and benefits, in this section we propose research questions in four broad areas.

\textbf{Understanding and mitigating risks} Most importantly, research should be conducted to understand whether and under what circumstances episodic memories can lead to unwanted behavior on the part of \textsc{ai} agents. This can be simultaneously complimented with investigations of how to prevent the undesirable behavior that episodic memory might enable. The creation of datasets with tasks designed to test the relative alignment of models with and without various forms of episodic memory would be useful.

Particularly initially, experiments could be done with episodic memories which are \textit{directly interpretable}, as discussed in Section VI(A). If memories are made up of natural language descriptions of events, possibly together with images from such events, it will be easier for researchers to understand the role that such memory representations play in shaping agent behavior.



\textbf{Using memories for safety}
We suggest two main directions for research into how memory can be used to enhance the safety of \textsc{ai} agents. First, methods could be developed to put into practice our proposals about using episodic memories for monitoring and control. In order to do so, many questions will have to be addressed, including the structure of the memories, how to enable arbitrary unanticipated queries of the memories, and what kinds of information is most important to store and present to users or other safety monitors. Because some aspects of memory representations might be designed or discovered during a learning process, it should be possible to influence these representations to make safety-relevant features of the remembered episodes easily retrievable.

Second, research will be needed to determine how to use episodic memories (or parts thereof) as components in an agent's planning process. Questions to be addressed here include determining what kind of training and guidance will be required for agents to learn to use the memories in a useful way; whether and how it is possible to make use of examples of undesirable actions in order to prevent similar actions in the future; and how to design a procedure that allows for certain kinds of newly acquired memories to influence future actions while disallowing others.

\textbf{Episodic memory architectures designed for safety}
We believe that how episodic memories are represented, stored, and accessed by an \textsc{ai} agent will soon be a popular and important field of research. How to do this without making agent behavior less safe, for example according to our proposed principles or others which may be developed, should be the focus of safety-minded efforts in this area. A significant question is whether there will be a trade-off between safety and usefulness according to some metric (e.g. accuracy, speed, memory use). Relatedly, how dissimilar to human memory in some respects (e.g. interpretability and controllability) can artificial episodic memory be while still enabling the range of functions that human memory does?

\textbf{Governance responses to episodic memory capabilities}
Episodic memory in \textsc{ai} agents is an important topic for researchers working on \textsc{ai} governance in two ways. First, memories might be employed as part of an \textsc{ai} governance strategy. How this might be implemented, in which domains might memories be most useful, and how memories can be used for governance purposes while preserving user privacy are some of the questions that will have to be addressed.

Second, if, as we have argued here, artificial episodic memories are a significant potential source of risk, work will have to be done to determine how governance efforts might best mitigate that risk. What kind of regulations or standards should govern agents' use of episodic memories? How would limitations or oversight differ when dealing with a new type of capability such as episodic memory rather than considering the amount of computing resources used or the size of a model?

\section{Conclusion}
We have examined some of the risks and benefits of episodic memory in \textsc{ai} agents. 
We note that there is some overlap between the sources of risks and benefits: positive aspects of episodic memory from the perspective of one of these may be negative when seen from the other. What might constitute a risk may also be seen as a benefit. For example, the ability to record, store, and recall information about what an agent does is beneficial for monitoring and control of the agent but poses risks to the privacy of the people interacting with the agent. How to balance these and other such trade-offs will benefit from broad and inclusive participation and investigation.

At the moment, many of the risks we warn about have not yet been seen in deployed models. Though we have been careful to ground our concerns in a discussion of the role of episodic memory in humans, some may view them as speculative. We contend, however, that the best time to begin considering the dangers of a new technology or technique is precisely when the risks are still open to speculation rather than already upon us. 

Developing the ability for \textsc{ai} agents to form, retrieve, and reason over episodic memories would introduce significant new capabilities and would represent a major milestone along the road to more advanced artificial intelligence. It is fortunate that these capabilities did not develop before concerns about \textsc{ai} reliability, safety, and alignment became more common within the \textsc{ai} research community. This presents the community with an opportunity to deliberatively and cautiously develop a potentially dangerous capability to ensure that it makes \textsc{ai} safer rather than more dangerous. We hope by bringing attention to this topic to foster a wider discussion of the risks and benefits of artificial episodic memory and contribute to the establishment of a research community to address them.

\bibliographystyle{IEEEtran}

\bibliography{memory_bib}

\end{document}